%% file: main.tex
\documentclass[conference]{IEEEtran}
\IEEEoverridecommandlockouts
\usepackage{cite}
\usepackage{xspace}
\usepackage{amsmath,amssymb,amsfonts}
\usepackage{multirow}
\usepackage{multicol}
\usepackage{graphicx}
\usepackage{textcomp}
\usepackage{xcolor}
\usepackage{colortbl}
\usepackage{subcaption} 
\usepackage{url}
\usepackage{booktabs}
\usepackage{algorithm}
\usepackage{algpseudocode}
\def\BibTeX{{\rm B\kern-.05em{\sc i\kern-.025em b}\kern-.08em
    T\kern-.1667em\lower.7ex\hbox{E}\kern-.125emX}}

\newcommand{\etal}{\textit{et al.}\xspace}

\begin{document}


\title{Progressive Retinal Image Registration via Global and Local Deformable Transformations\\
}


\author{
\IEEEauthorblockN{1\textsuperscript{st} Yepeng Liu}
\IEEEauthorblockA{\textit{School of Computer Science} \\
\textit{Wuhan University}\\
Wuhan, China \\
yepeng.liu@whu.edu.cn}
\and
\IEEEauthorblockN{2\textsuperscript{nd} Baosheng Yu}
\IEEEauthorblockA{\textit{School of Computer Science} \\
\textit{The University of Sydney}\\
Sydney, Australia \\
baosheng.yu@sydney.edu.au}
\and
\IEEEauthorblockN{3\textsuperscript{rd} Tian Chen}
\IEEEauthorblockA{\textit{School of Computer Science} \\
\textit{Wuhan University}\\
Wuhan, China \\
tian.chen@whu.edu.cn}
\and
\IEEEauthorblockN{4\textsuperscript{th} Yuliang Gu}
\IEEEauthorblockA{\textit{School of Computer Science} \\
\textit{Wuhan University}\\
Wuhan, China \\
yuliang\_gu@whu.edu.cn
}
\and
\IEEEauthorblockN{5\textsuperscript{th} Bo Du}
\IEEEauthorblockA{\textit{School of Computer Science} \\
\textit{Wuhan University}\\
Wuhan, China \\
dubo@whu.edu.cn
}
\and
\IEEEauthorblockN{6\textsuperscript{th} Yongchao Xu*}
\IEEEauthorblockA{\textit{School of Computer Science} \\
\textit{Wuhan University}\\
Wuhan, China \\
yongchao.xu@whu.edu.cn
}
\and
\IEEEauthorblockN{7\textsuperscript{th} Jun Cheng}
\IEEEauthorblockA{\textit{Institute for Infocomm Research} \\
\textit{A*STAR}\\
Singapore \\
cheng\_jun@i2r.a-star.edu.sg}
\thanks{* Corresponding Author}
}

\maketitle

\begin{abstract}
Retinal image registration plays an important role in the ophthalmological diagnosis process. Since there exist variances in viewing angles and anatomical structures across different retinal images, keypoint-based approaches become the mainstream methods for retinal image registration thanks to their robustness and low latency. These methods typically assume the retinal surfaces are planar, and adopt feature matching to obtain the homography matrix that represents the global transformation between images. Yet, such a planar hypothesis inevitably introduces registration errors since retinal surface is approximately curved. This limitation is more prominent
when registering image pairs with significant differences in viewing angles. To address this problem, we propose a hybrid registration framework called HybridRetina, which progressively registers retinal images with global and local deformable transformations. For that, we use a keypoint detector and a deformation network called GAMorph to estimate the global transformation and local deformable transformation, respectively.
Specifically, we integrate multi-level pixel relation knowledge to guide the training of GAMorph. 
Additionally, we utilize an edge attention module that  includes the geometric priors of the images, ensuring  the deformation field focuses more on the vascular regions of clinical interest.
Experiments on two widely-used datasets, FIRE and FLoRI21, show that our proposed HybridRetina significantly outperforms some state-of-the-art methods. The code is available at https://github.com/lyp-deeplearning/awesome-retinal-registration.

\end{abstract}


\section{Introduction}
\label{sec:intro}



\begin{figure}[t]
    \centering
    \begin{subfigure}[b]{\linewidth}
        \centering
        \includegraphics[width=0.90\linewidth]{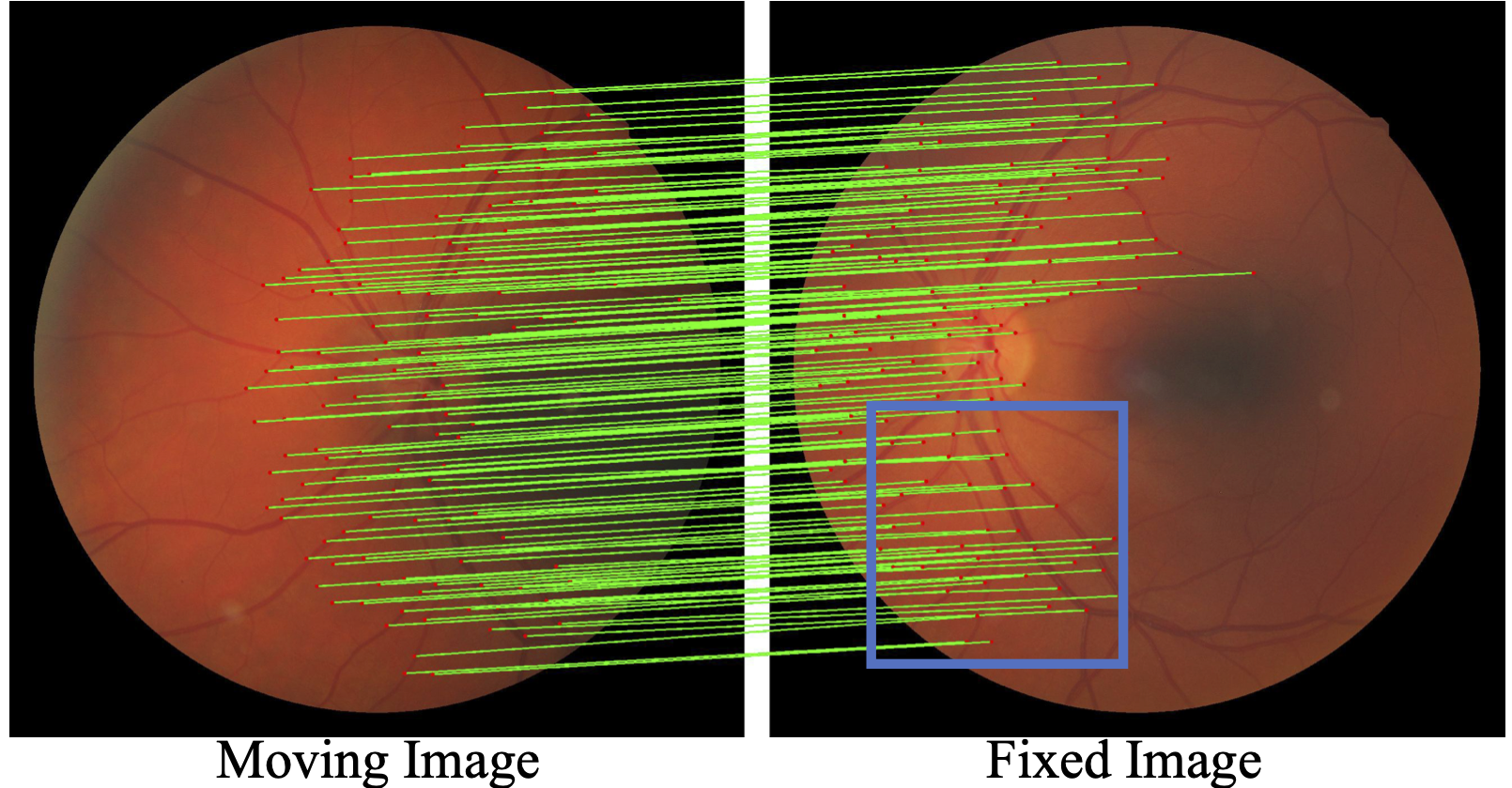}
        \vspace{-3pt} 
        \caption{Feature matching result in global transformation by SuperRetina}
        \label{fig:image1}
    \end{subfigure}

    \begin{subfigure}[b]{\linewidth}
        \centering
        \includegraphics[width=0.90\linewidth]{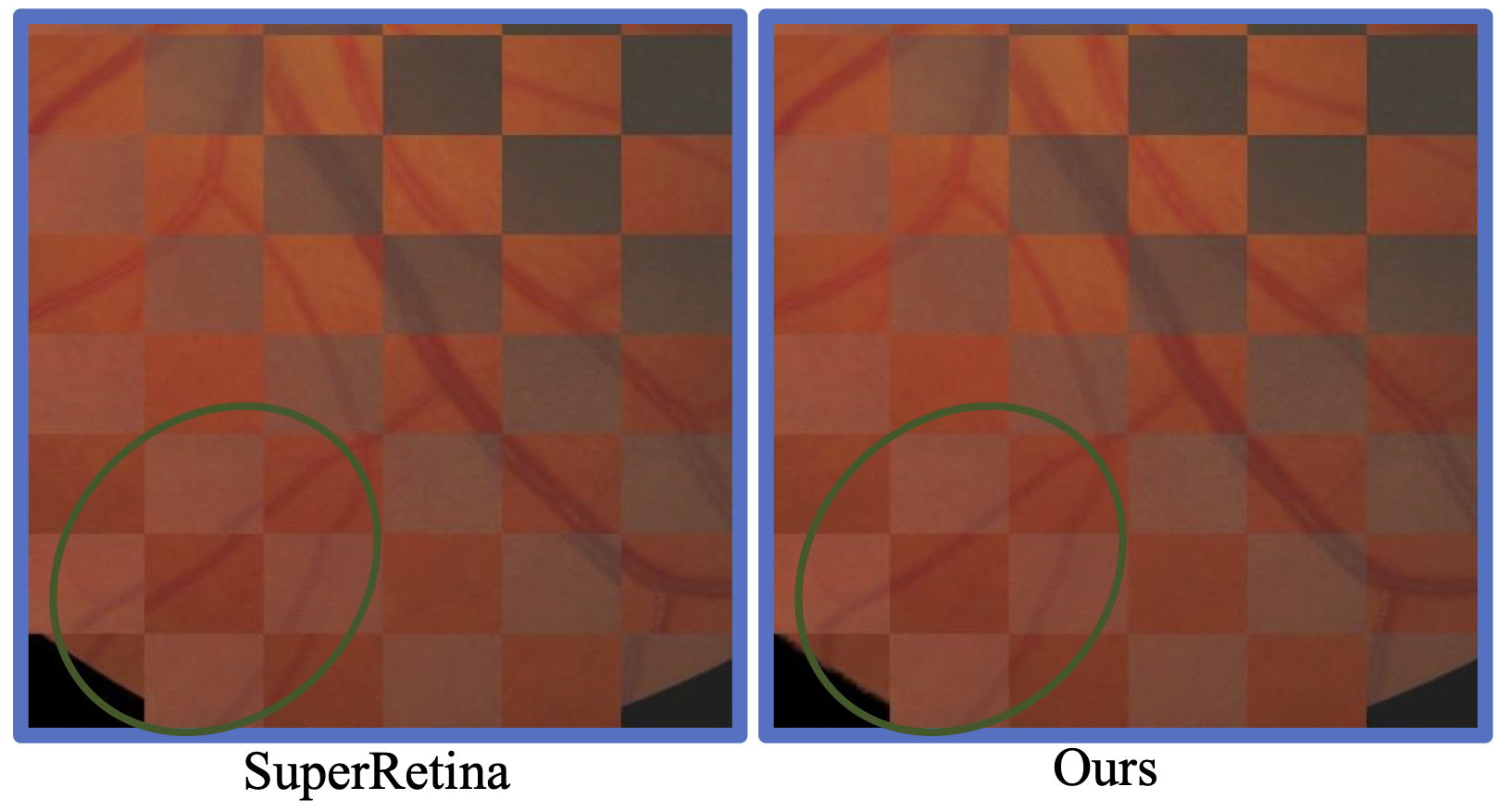}
        \vspace{-3pt} 
        \caption{ The registration results of SuperRetina and our proposed method}
        \label{fig:image2}
    \end{subfigure}

    \caption{
    An example demonstrates the registration results of the global transformation  and our progressive transformation.
    We use mosaic images to compare the registration performance, and the continuity of blood vessels across tiles reflects the registration quality.
    SuperRetina~\cite{liu2022semi} exhibits excellent feature matching performance in global transformation, but the registration results still contain obvious errors. 
}
    \label{fig:motivation}
\end{figure}

According to the World Health Organization, approximately 220 million people worldwide suffer from various degrees of visual impairments, with around 110 million individuals afflicted by preventable eye diseases that have gone untreated~\cite{motta2019vessel}. Retinal image registration aims to align multiple images of the same retina obtained from different perspectives and periods. This technique has been substantiated as crucial in medical diagnostics as it enables clinicians to track changes in the retina over time and facilitates precise laser treatment for conditions such as diabetic retinopathy and age-related macular degeneration~\cite{truong2019glampoints, liu2022semi}. 

Retinal image registration methods can be roughly grouped into intensity-based and keypoint-based. Intensity-based registration algorithms rely on optimization algorithms to find appropriate matching relationships based on pixel intensities, and are often time-consuming at high resolutions~\cite{chen2010partial, zaslavskiy2008path, ritter1999registration,oinonen2010identity}. 
With advancements in deep learning, keypoint-based methods have rapidly progressed by learning effective feature representations~\cite{detone2018superpoint,revaud2019r2d2,liumos,liu2022semi}, as early hand-crafted features struggle to address variations in lighting and anatomy within retinal images~\cite{david2004distinctive, bay2006surf, cattin2006retina}.

The keypoint-based methods typically employ keypoint detection and description to find homography transformations between different images. 
SuperRetina~\cite{liu2022semi} achieves impressive registration performance in retinal image registration by using a semi-supervised training scheme to simultaneously predict keypoints and descriptors.
However, keypoint-based methods inherently introduce registration errors due to approximating the retinal surface as planar to calculate the homography transformation.
As depicted in Fig.~\ref{fig:motivation},  SuperRetina~\cite{liu2022semi} achieves very good feature matching results, but its warped moving image still exhibit many mismatches with the fixed image. 
Some recent works~\cite{ding2022combining,zhang2021two} attempt to obtain vessel masks through additional vessel segmentation and then use nonlinear optimization algorithms to align local regions. However, these approaches introduce additional mask annotation costs and inference time. Directly performing nonlinear alignment on original color fundus images is challenging due to the presence of various background elements outside the blood vessels.

To address this problem, we propose a novel progressive retinal image registration framework called HybridRetina. In the global registration stage, we use a keypoint detector to extract keypoints and calculate the homography matrix, obtaining the coarsely matched image pair. 
Subsequently, we employ a geometry-aware deformation network, GAMorph,  to perform nonlinear alignment for the coarsely matched image pair in the local deformable registration stage. 
Moreover, we use the \textbf{multi-level pixel relation} priors obtained from the global matching stage to guide the movement of pixels in the deformation field.
Additionally, we propose an \textbf{edge attention module}  to ensure that the deformation field focuses more on the alignment of vascular regions.
Notably, our deformation field is trained on four open-source color fundus image datasets and tested on both color fundus images and fluorescein angiography images.
The results demonstrate the generalization and effectiveness of HybridRetina, outperforming some state-of-the-art performance in retinal image registration.

\begin{figure*}[!ht]
    \centering
\includegraphics[width=0.95\linewidth]{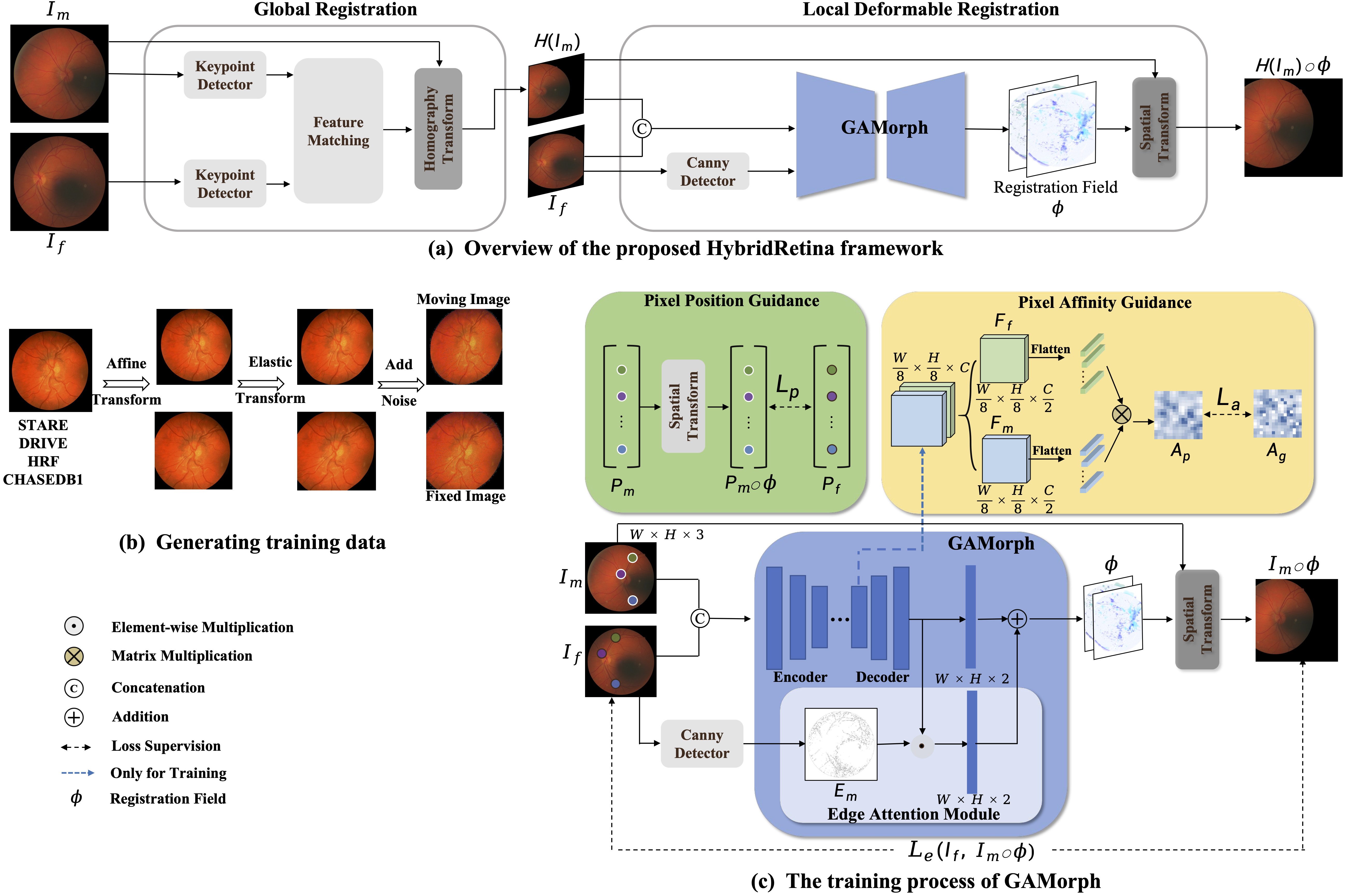}
    \caption{
    Illustration of Our HybridRetina Framework. 
(a) The steps of global registration and local deformable registration.
(b) The process of generating paired data for our proposed GAMorph.
(c) The training process of GAMorph. 
    }
\label{fig:network_arth}
\end{figure*}

\section{Related Work}
\label{sec:relatedwork}
\subsection{Global Image Registration}
Keypoint-based global registration algorithms typically involve four steps: detection of interest points, computing feature descriptors, feature matching, and estimation of homographic transformation between the images. 
Many research  focus on how to extract more replicable and reliable keypoints~\cite{potje2023enhancing, bay2006surf, detone2018superpoint, revaud2019r2d2}. SIFT~\cite{david2004distinctive} computes corners and blobs on different scales to achieve scale invariance and extracts descriptors using the local gradients. SuperPoint~\cite{detone2018superpoint} provides a self-supervised approach to simultaneously train keypoint detection and description. R2D2~\cite{revaud2019r2d2} learns keypoint detection and description jointly with prediction of the local descriptor discriminativeness to  find reliable keypoints.


\subsection{Deformable Image Registration}
In the past, there have been numerous non-deep learning nonrigid registration algorithms. Several studies optimize within the space of displacement vector fields. Daniel \etal ~\cite{rueckert1999nonrigid} use affine transformations for global motion matching and use free-form deformation based by B-splines to describe local variations. 
Tom \etal ~\cite{vercauteren2009diffeomorphic} adapt the optimization procedure underlying the Demons algorithm to a space of diffeomorphic transformations.
With the rise of deep learning techniques, there has been an emergence of registration algorithms based on  deep learning~\cite{ balakrishnan2019voxelmorph,chen2022transmorph,shi2022xmorpher,fan2023automated,wang2024recursive, chen2023transmatch}. The initial algorithms were trained based on supervised labels~\cite{sokooti2017nonrigid, yang2017quicksilver}. However, these supervised algorithms typically require deformation fields as ground truth. Subsequently, unsupervised deformation fields gained prominence. Voxelmorph~\cite{balakrishnan2019voxelmorph} rapidly computes a deformation field by unsupervised training from coarsely aligned data. 
TransMorph~\cite{chen2022transmorph} uses a hybrid Transformer-ConvNet model and achieves better registration results on image pairs with substantial deformations. 
Although these  unsupervised methods such as VoxelMorph and TransMorph have shown to be promising for image registration, it is challenging to apply them to retinal images, as retinal image registration requires precise alignment in many regions such as the vessel area.

\subsection{Retina Image Registration}
Most of existing retinal image registration methods are based on rigid transformation~\cite{deng2010retinal, braun2018eyeslam,truong2019glampoints}.   LoSPA ~\cite{addison2015low} and DeepSPA~\cite{lee2019deep} focus on describing image patches by step pattern analysis for homography estimation.  REMPE~\cite{hernandez2020rempe} first finds many candidate points by vessel bifurcation detection and the SIFT detector~\cite{david2004distinctive}, and then performs point pattern matching   based on eye modelling and camera pose estimation to identify geometrically valid matches. 
SuperRetina~\cite{liu2022semi} proposes progressive keypoint expansion to enrich the keypoint labels at each training epoch. By utilizing a keypoint-based improved triplet loss as its description loss, SuperRetina produces highly discriminative descriptors. GeoFormer~\cite{liu2023geometrized} is trained in a fully self-supervised manner by minimizing a multi-scale cross entropy based matching loss on auto-generated training data. 

In this paper, we propose to integrate rigid registration with deformable models for retinal image registration.

\section{Method}
\label{sec:work}

The keypoint-based algorithms are unable to handle the geometric distortions that occur when mapping the nearly spherical retinal surface onto the image plane. To address this issue, we propose a progressive retinal image registration framework. 
In this section, we first introduce the progressive registration process (Sec.~\ref{hrp}). Next, we introduce our proposed multi-level pixel relation guidance (Sec.~\ref{guidance}) and edge attention module (Sec.~\ref{edge}), which assist in the training of our deformation field prediction network, GAMorph. Finally, we introduce the loss function of GAMorph (Sec.~\ref{loss_method}).

\subsection{Progressive Registration} 
\label{hrp}
Fig.~\ref{fig:network_arth}.a illustrates the process of our progressive registration framework. Given the source moving image $I_m \in \mathbb{R}^{W \times H \times 3}$ and the fixed target image $I_f \in \mathbb{R}^{W \times H \times 3}$, we first use a keypoint detector to extract keypoints and corresponding descriptors. 
Descriptors are matched using the nearest neighbor algorithm in feature matching stage.
Subsequently, the RANSAC algorithm is utilized to obtain the global homography transformation matrix between the image pair, transforming the moving image $I_m$ into $H(I_m)$. The output of the global registration is fed into the network GAMorph to predict the registration field $\phi\in \mathbb{R}^{W \times H \times 2}$. Finally, we utilize the spatial transform operation \cite{jaderberg2015spatial} to get the warped moving image $I_m \circ \phi$.

We use the pre-trained SuperRetina~\cite{liu2022semi} network as the keypoint detector to provide the keypoints and descriptors needed for global feature matching in the training stage. It is noteworthy that during the inference phase, we can use any keypoint detector to replace SuperRetina~\cite{liu2022semi}. 
The deformation field network GAMorph is based on the UNet architecture, which includes a four-layer encoder and a four-layer decoder to extract features. Then, a prediction head composed of a 3x3 convolutional layer is used to predict the deformation field $\phi$. Another branch utilizes the  edge attention module, incorporating the geometric priors of the image edges and combining with the deformation field output from the prediction head through an add operation.



\subsection{Multi-level Pixel Relation Guidance}
\label{guidance}
The  vanilla deformable registration networks~\cite{balakrishnan2019voxelmorph, chen2022transmorph} typically assume an equal relationship between pixels.
Consequently, they yield suboptimal deformation fields that tend to overfit specific local regions, particularly complex backgrounds. 
Fortunately, the feature matching results in the global registration stage provides some valuable gifts. In this subsection, we utilize these gifts to establish pixel relationship priors, thereby better supervising the learning of the deformation field.

\subsubsection{Pixel Affinity Guidance (PAG)}
\label{affinity}
\begin{figure}[t]
    \centering
\includegraphics[width=1.0\linewidth]{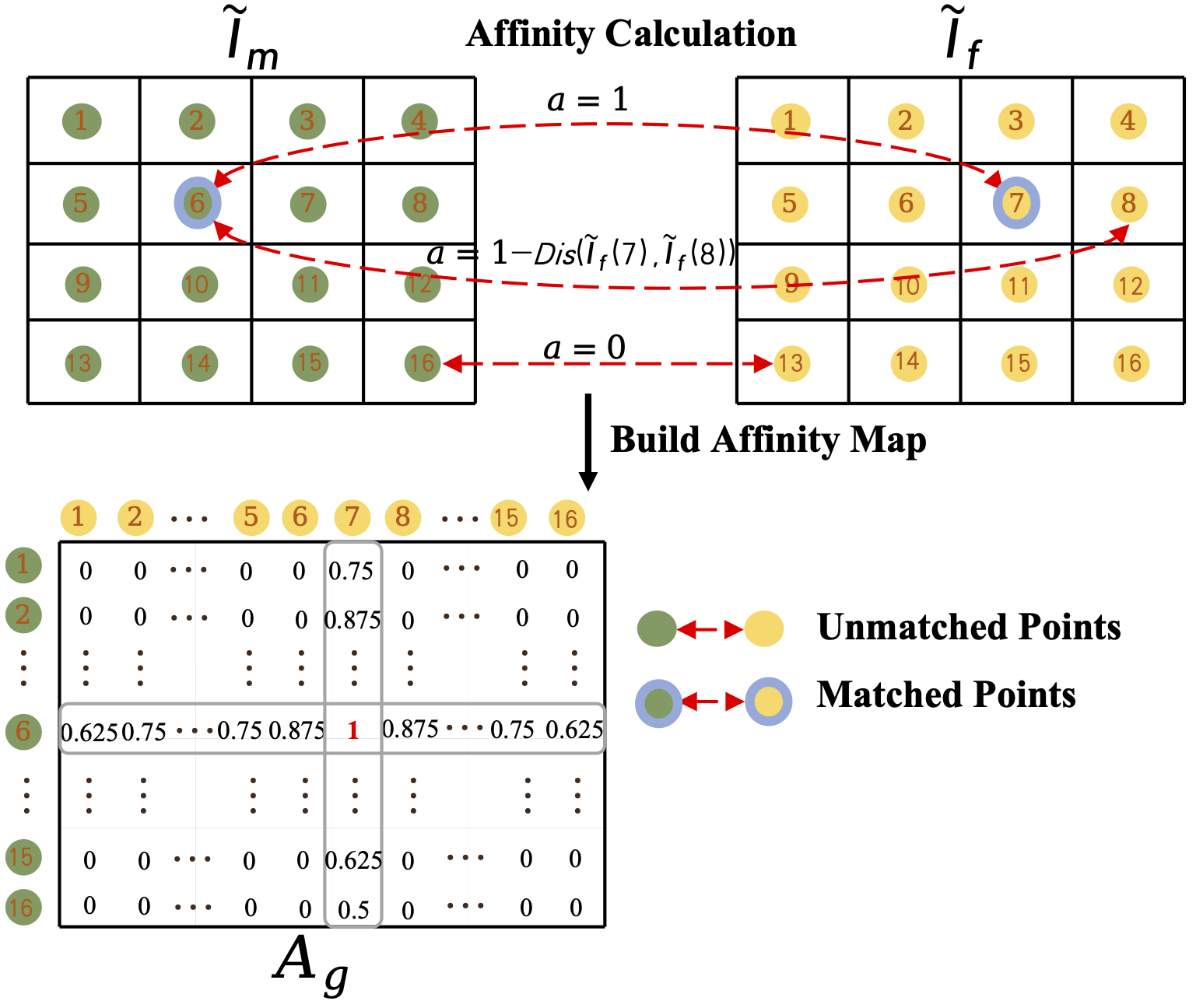}
    \caption{A toy illustration of computing affinity map $A_g$. We use matching prior and Manhattan distance to calculate the affinity value $a$ between all pixels.}
\label{fig:affinity1}
\end{figure}

The registration field aims to align all pixels of the moving image to the fixed image. 
We use affinity to describe the degree of association between pixels in the image pair. 
We observe that  matched keypoints information provided by  the global registration stage offers a pixel relationship prior for the deformation field. 
These discrete matched keypoints should have a strong affinity value. Additionally, due to the properties of diffeomorphic~\cite{balakrishnan2019voxelmorph}, pixels in the neighborhood of the matched keypoints are also likely to move to the vicinity of the corresponding matched points. Based on these priors, we leverage both the matching information and the Manhattan distance to construct an affinity map $A_g$ between all pixels of the moving image and the fixed image.

As shown in Fig.~\ref{fig:affinity1}, let $\Tilde{I}_{m} \in \mathbb{R}^{W'\times H'\times 3}$ and $\Tilde{I}_{f} \in \mathbb{R}^{W'\times H'\times 3}$ denote the downsampled moving and fixed images, respectively. In our experiments, we use the downsampling stride of 8, \textit{i.e.}. The ground-truth affinity map $A_g \in \mathbb{R}^{H'W'\times H'W'}$ is then constructed to compute the relationships among all pixels. Let the pixel points in $\Tilde{I}_{m}$ and $\Tilde{I}_{f}$ be represented by \( (x_i, y_i) \) and \( (x_j, y_j) \) , respectively. The elements \( A_g(i,j) \) of the affinity relationship matrix \( A_g \) are computed as follows:

{\scriptsize{
\begin{equation}
A_g(i,j) = 
\begin{cases} 
1, & \text{if } (x_i, y_i)  \text{ matches } (x_j, y_j)  \\
1-\| (x_j, y_j) - (x_m, y_m) \|_1, & \text{if } (x_i, y_i)  \text{ matches } (x_m, y_m) \text{ in } \Tilde{I}_{f} \\
1-\| (x_n, y_n) - (x_i, y_i) \|_1, & \text{if } (x_n, y_n) \text{ in } \Tilde{I}_{m} \text{ matches } (x_j, y_j)  \\
0, & \text{otherwise}
\end{cases},
\label{eq:relation value}
\end{equation}
}
}
where \(\| \cdot \|_1\) indicates the normalized Manhattan distance.

To obtain the predicted affinity map, we employ an affinity prediction branch that exists only during the training phase as depicted in Fig.~\ref{fig:network_arth}.c. Specifically, we directly split  the feature map obtained from the decoder into $F_m$ and $F_f$, representing the pixels of moving image and fixed image, respectively.
Then, we flatten the feature maps $F_m$ and $F_f$ and conduct matrix multiplication, \textit{i.e.}, the predicted affinity map is computed as $A_p = F_m F_f^{\top} \in \mathbb{R}^{H'W' \times H'W'}$. Lastly, we evaluate the loss between the predicted and the ground-truth affinity maps as follows:
\begin{equation}
    L_a = \mathcal{L}_{mse}(A_p,~A_g).
\end{equation}

\subsubsection{Pixel Position Guidance (PPG)}
\label{point}
The position information of the matched keypoint pairs can explicitly supervise the local predictions of the deformation field. However, it should be noted that the keypoint positions from the keypoint detector may not be entirely accurate, as the detector uses Non-Maximum Suppression (NMS) to identify local maximum response points. We observe that keypoints with prominent geometric features in an image are often more reliable, such as points on blood vessels. Therefore, we calculate the sum of horizontal and vertical gradients for keypoints on the reference image and select the top 120 keypoints as the supervisory signal.
Let $P_m \in \mathbb{R}^{N\times 2}$ and $P_f \in \mathbb{R}^{N\times  2}$ denote the selected keypoints from the moving image and the fixed image, respectively. The loss for local pixel position guidance can be defined as:
\begin{equation}
L_{p} = \mathcal{L}(P_m \circ \phi, ~P_f),
\label{eq:lreg loss}
\end{equation}
where $\mathcal{L}$ represents the smooth-$\ell_1$ loss function defined in~\cite{ren2015faster}, and  $P_m \circ \phi$ represents the corresponding keypoints in warped moving image obtained by the registration field $\phi$.

\begin{figure}[t]
    \centering
\includegraphics[width=1.0\linewidth]{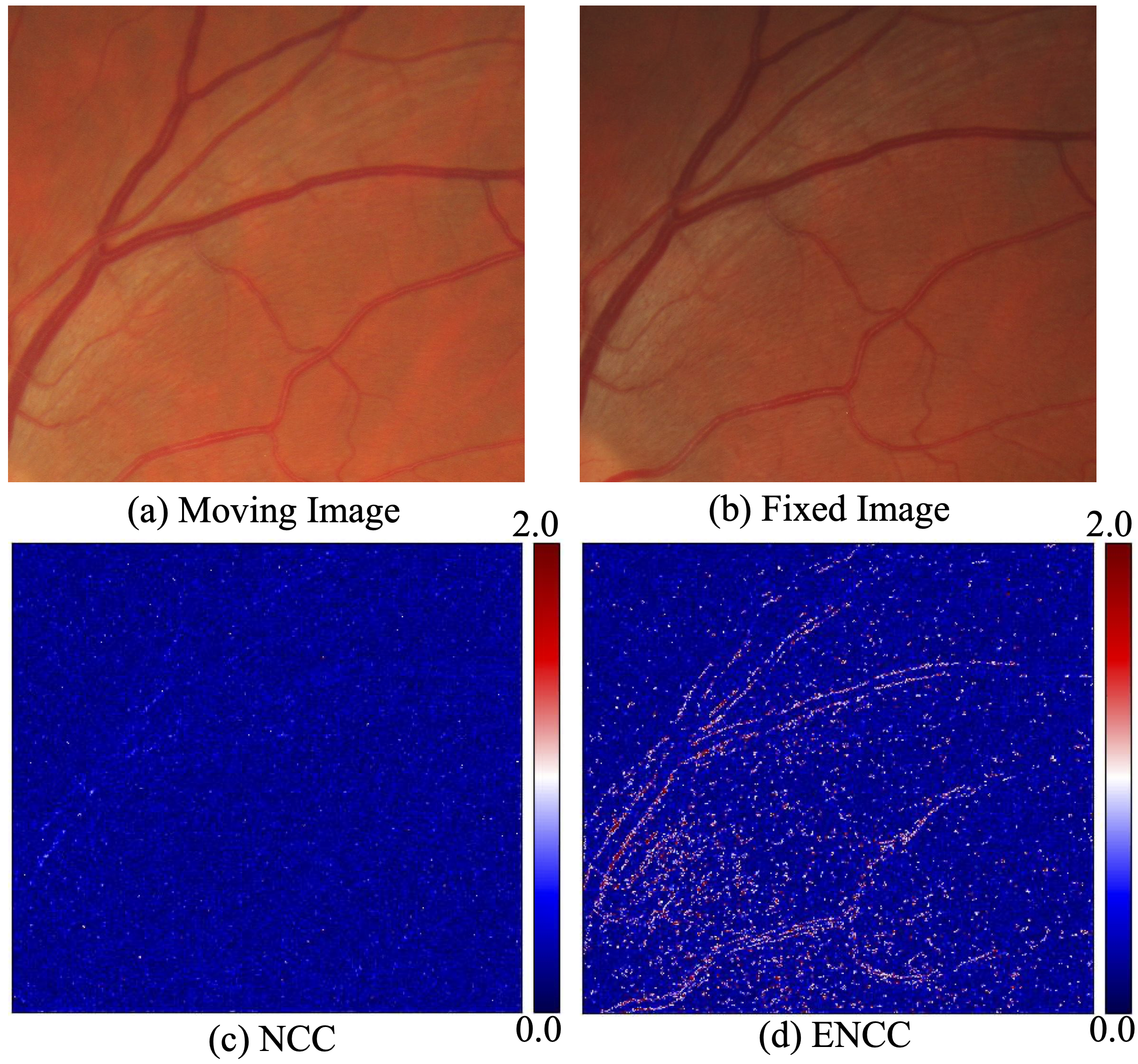}
    \caption{Comparison of pixel-wise differences between the moving image and the fixed image using different similarity measurement methods. Results of normalized cross-correlation (NCC) and our proposed edge-guided normalized cross-correlation (ENCC) are shown.}
\label{fig:attention_loss}
\end{figure}

\subsection{Edge Attention Module}
\label{edge}

Learning deformation fields involves minimizing pixel-wise differences while preserving diffeomorphic properties. However, commonly used similarity loss functions such as   normalized cross-correlation (NCC) \cite{balakrishnan2019voxelmorph}   tend to prioritize global similarity at the expense of local fine-grained alignment. 
As shown in Fig.~\ref{fig:network_arth}.c, we propose an edge attention module to emphasize the importance of edge priors for improved local fine-grained alignment. 
Specifically, we first employ the Canny detector~\cite{canny1986computational} with adaptive  thresholds to extract edges  from the fixed image.
Then, we derive a binary edge mask $E_m$ and add it to the output of the decoder to obtain the final deformation field that includes edge priors. In addition, we propose a new edge-guided NCC (ENCC) loss  $L_e$ by integrating edge mask with the original NCC loss \cite{balakrishnan2019voxelmorph}.

As shown in Fig.~\ref{fig:attention_loss}, our proposed ENCC loss focuses more on changes in blood vessels and assigns greater weight to the edge response regions, making it more similar to clinical diagnosis. 
For training, we need to minimize the error between the fixed image $I_f$ and the warped moving image $I_t=(I_m \circ \phi)$. The loss function $L_e$ is calculated as follows:

{\scriptsize{
\begin{equation}
L_{e} = \sum_{p} \frac{\left(1 + E_m(p)\right)\left(\sum_{p_i \in \Omega_p}\left(I_f(p_i)-\bar{I}_f(p)\right)\left(I_t(p_i)-\bar{I}_t(p)\right)\right)^2}{\left(\sum_{p_i \in \Omega_p}\left(I_f(p_i)-\bar{I}_f(p)\right)^2\right)\left(\sum_{p_i \in \Omega_p}\left(I_t(p_i)-\bar{I}_t(p)\right)^2\right)},
\end{equation}
}
}
where $\Omega_p$ denotes the set of neighboring points near point $p$, and $\bar{I}_f$ and $\bar{I}_t$ denotes mean values within local $w\times w$ ($w=15$) window centered at $p$.

\input{tables/table_fire_benchmark}

\subsection{Overall Loss}
\label{loss_method}


As shown in  Fig.~\ref{fig:network_arth}.b, we  use the open-source color fundus retinal images to construct the training image pairs. 
The multi-level pixel relation guidance module generates the losses 
$L_a$ and $L_p$. Meanwhile, the similarity loss $L_e$ is obtained by applying the ENCC loss function to the warped moving image and the fixed image.
Finally,  we also incorporate regularization terms $L_{smooth}$ inspired by VoxelMorph~\cite{balakrishnan2019voxelmorph}, which enforces a spatially smooth deformation as a function of the spatial gradients of registration field. The overall loss function is as follows:
\begin{equation}
  L_{all} = L_{a}+ \lambda_1 L_{p} + \lambda_2 L_{e}  +  \lambda_3 L_{smooth},
  \label{eq:lall loss}
\end{equation}
where $\lambda_1$, $\lambda_2$, and $\lambda_3$ are employed to balance different loss terms. In our experiments, we empirically set $\lambda_1=5.0$, $\lambda_2=2.0$, and $\lambda_3=1.0$ to ensure that all loss terms exhibit similar gradient norms.

\section{Experiments}

\subsection{Datasets and Evaluation Metrics}
Our model is trained on four public color fundus datasets and tested on two challenging retinal registration datasets.

\noindent\textbf{The Training dataset:} We utilize four publicly available retinal color fundus datasets, including STARE \cite{hoover2003locating}, DRIVE \cite{staal2004ridge}, CHASEDB1 \cite{fraz2012ensemble}, and HRF \cite{odstrvcilik2009improvement}, totaling 768 images without annotated keypoints.

\noindent\textbf{The FIRE  dataset:}\footnote{\url{https://projects.ics.forth.gr/cvrl/fire/}}~\cite{hernandez2017fire} The first public test set we used is the FIRE dataset, which has 134 registered color fundus image pairs of size $2912 \times 2912$ acquired by a Nidek AFC210 fundus camera.
The images are split into three categories with different characteristics:
1) Category-$\mathcal{S}$ contains 71 pairs with large overlaps and no anatomical changes, making them easy to register; 2) Category-$\mathcal{A}$ contains 14 pairs with large overlaps and significant anatomical changes, making them moderately hard to register; and 3) Category-$\mathcal{P}$ contains 49 pairs with small overlaps, making them hard to register. Each pair of images is manually annotated with 10 pairs of points, primarily located at vessel junctions and intersections. 

\noindent\textbf{The FLORI21  dataset:}\footnote{\url{https://ieee-dataport.org/open-access/}}~\cite{ding2021flori21} The second public test set we used is the FLORI21 dataset, which provides 15 pairs of ultra-widefield (UWF) fluorescein angiography images. The fluorescein angiography images have a large field of view (FOV) up to $200^{\circ}$ compared to narrow-field images (between $30^{\circ}$ and $50^{\circ}$). 
Each pair of images is annotated with 15 pairs of points at vessel birucations, ensuring coverage of the entire image area.

\noindent\textbf{Evaluation Metrics.}
Each image pair is annotated with at least 10 pairs of points  distributed globally across the image. Following ~\cite{liu2022semi}, we measure the registration performance based on the root mean square error (RMSE) , acceptable rate (Accept) and area under curve (AUC) metrics. 

Let $K_f$ denote the ground truth points on the fixed image, and $K_m'$ represent the ground truth points on the moving image after being transformed by the registration algorithm. We first calculate the RMSE metric between $K_f$ and $K_m'$. 
Next, we calculate the acceptable rate for all images, where a case is considered successful if  RMSE $<$ 20.
Finally, we calculate the AUC metric~\cite{hernandez2017fire}, which requires computing the expectation of success rate at different threshold values of RMSE. 

\subsection{Implementation Details}

We implement the proposed GAMorph using PyTorch. The models are trained on a single NVIDIA GeForce RTX 2080 Ti GPU, and all input images are resized to $768\times 768$ pixels. 
We employ the extensively utilized Adam optimizer, setting a learning rate of 0.001 and a batch size of 8. To generate paired training images, we apply random affine transformations, including rotation, translation, and scaling. Additionally, we utilize elastic transformations~\cite{simard2003best} to introduce deformations in the images, specifically using an intensity value of $50$ and a standard deviation of $6$ for the Gaussian filter kernel to ensure appropriate elastic deformation. Finally, we add Gaussian noise to increase the differences between pixels.

\begin{figure*}[t]
    \centering
\includegraphics[width=0.9\linewidth]{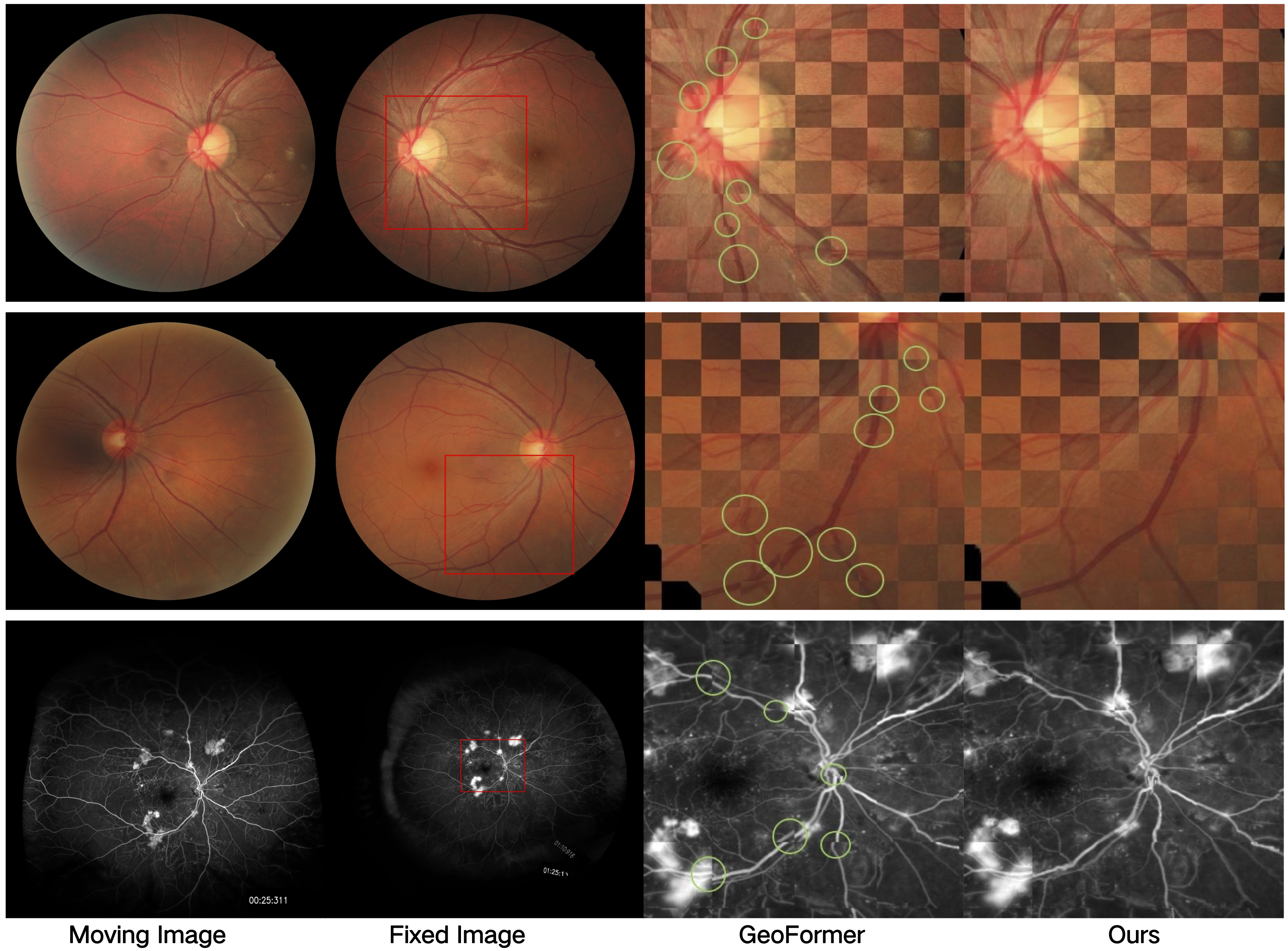}
    \caption{Mosaic visualization of registration results. We show the alignment in the region enclosed by the red box. We observe misalignment in the  results by GeoFormer~\cite{liu2023geometrized}, highlighted by green circles. Best viewed electronically.}
    \label{fig:main_vis}
\end{figure*}

\subsection{Performance Results on FIRE}

We mainly compare with two types of state-of-the-art methods on the FIRE dataset:
\begin{itemize}
    \item Traditional algorithms: SIFT~\cite{david2004distinctive}, PBO~\cite{oinonen2010identity} and REMPE~\cite{hernandez2020rempe}.
    \item Deep learning-based methods: GLAMPoints~\cite{truong2019glampoints}, SuperGlue~\cite{sarlin2020superglue}, R2D2~\cite{revaud2019r2d2}, NCNet~\cite{rocco2020ncnet}, SuperRetina~\cite{liu2022semi}, and GeoFormer~\cite{liu2023geometrized} .
\end{itemize}

In Tab.~\ref{tab:fire_results}, we demonstrate the results of our HybridRetina framework using four different keypoint detectors.
It can be observed that our proposed GAMorph improves these baselines on mAUC metric by 6.0\%, 5.3\%, 4.8\% and 5.3\%, respectively. 
It is noteworthy that GAMorph only uses SuperRetina as the keypoint detector during training, and does not require fine-tuning when paired with other keypoint detectors.

Our HybridRetina achieves the highest mAUC of 80.9\% when using GeoFormer~\cite{liu2022semi} as the keypoint detector. 
Compared with the best traditional method REMPE~\cite{hernandez2020rempe}, we achieve the same AUC in the easy subset $\mathcal{S}$. However, REMPE struggles with the $\mathcal{A}$ subset, which has significant anatomical changes, and the $\mathcal{P}$ subset, which has small overlap.
Moreover, REMPE requires three minutes to register each image pair,  whereas our GAMorph only requires 32 milliseconds.
It is noteworthy that our algorithm achieves the best overall performance in both  challenging subsets, $\mathcal{A}$ and $\mathcal{P}$.
Visualized checkerboard images in Fig.~\ref{fig:main_vis} highlight our method's superior alignment compared to GeoFormer~\cite{liu2023geometrized}.

\input{tables/flori_result}

\subsection{Performance Results on FLORI21}
The FLORI21 dataset presents challenges due to significant scale differences and small overlaps between images. 
Additionally, our deformation network GAMorph is trained using only color fundus images, and it has not seen  fluorescein angiography images.
As shown in Tab.\ref{tab:flori_test}, our method improves the AUC by 5.9\%, 6.5\%, 7.2\%, and 6.6\% for the four global registration detectors, respectively.

\input{tables/ablation_component}

\subsection{Ablation Studies}
To justify the effectiveness and robustness of our proposed geometry guidance, we conduct a set of ablation studies on the FIRE dataset and FLoRI21 dataset. In these experiments, we utilize SuperRetina as the keypoint detector.

\noindent\textbf{Ablation study on key components of GAMorph.} We perform ablation study to evaluate each individual module's influence. 
Our baseline includes only the NCC loss function and the regularization term $L_{smooth}$.
As dedicated in Tab.~\ref{tab:ablation_componet}, all the three components significantly contribute to the HybridRetina. Pixel affinity guidance (PAG) improves the AUC by 1.2\% and 1.7\%, respectively. Pixel position guidance (PPG) improves the AUC by 1.6\% and 2.5\%, respectively.
Edge attention module (EAM) and  the ENCC loss function together improve the AUC by 1.6\% and 2.2\%, respectively.

\input{tables/ablation_deformable}
\noindent\textbf{Ablation study on using different deformable methods.}
We compare our GAMorph with other traditional and deep learning-based deformable methods.
To ensure a fair comparison, we retrained deep learning networks VoxelMorph~\cite{balakrishnan2019voxelmorph}, AC-RegNet~\cite{mansilla2020learning}, and TransMorph~\cite{chen2022transmorph} on our training dataset. As shown in Tab.~\ref{tab:ablation_deform}, our proposed GAMorph achieves the best performance, outperforming TransMorph by 2.9\% on the FIRE dataset and by 3.1\% on the FLORI21 dataset.



\subsection{Runtime Analysis}
The run time of network is computed over 134 pairs of images with a resolution of $768 \times 768$ pixels. 
We validate the efficiency of our algorithm on an NVIDIA GeForce RTX 3090 GPU. Our HybridRetina framework, combining SuperRetina and GAMorph, achieves an average processing time of 85ms per image, with GAMorph only requiring 32ms.

\section{Conclusion}

Retinal image registration typically relies on a global homography transformation using keypoints and their descriptions. To further improve the accuracy of global transformation methods, we propose a progressive  registration framework that includes both global registration and local deformable registration. Our method significantly improves the performance of state-of-the-art algorithms on two widely used datasets.
In the future, we would like to explore the potential of our method in multimodal medical image registration.

\section*{Acknowledgment}
This work was supported in part by the National Key Research and Development Program of China (2023YFC2705700), NSFC 62222112, 62225113, and 62176186, the Innovative Research Group Project of Hubei Province under Grants (2024AFA017).

\bibliographystyle{IEEEtran}
\bibliography{main.bib}
\end{document}

%% file: tables/table_fire_benchmark.tex
\begin{table*}[t] 
   \centering
       \caption{Quantitative comparison with state-of-the-art methods for color fundus image registration on the FIRE dataset~\cite{hernandez2017fire}. We present the results of GAMorph with different keypoint detectors. The performance improvements of our method over the baseline are marked in blue.
       }
     \begin{tabular}{l|c|c|cccc}
     \toprule
       Method &  RMSE$\downarrow$  &  Accept [\%]$\uparrow$ &{{AUC-$\mathcal{S}$} [\%]$\uparrow$} & {{AUC-$\mathcal{A}$} [\%]$\uparrow$} &{{AUC-$\mathcal{P}$} [\%]$\uparrow$} & {{mAUC} [\%]$\uparrow$} \\
      \midrule
      PBO, ICIP2010 \cite{oinonen2010identity}  & 18.6 &70.9 &84.4 & 69.1 & 12.2 & 55.2   \\
      REMPE, JBHI2020 \cite{hernandez2020rempe}  & 6.3 &97.0 &95.8 & 66.0 & {54.2} & 72.0   \\
      GLAMpoints, ICCV2019 \cite{truong2019glampoints} & 13.6 &92.5 & 85.0 & 54.3 & 47.4  & 62.2  \\    
      SuperGlue, CVPR2020 \cite{sarlin2020superglue} & 9.4 &95.5 & 88.5 & 68.9 & 48.8 & 68.7  \\
      R2D2, NIPS2019 \cite{revaud2019r2d2} & 6.4 & 95.5 &92.8 & 66.6 & 54.0& 71.1 \\
      NCNet, TPAMI2022 \cite{liu2022semi} & 14.1 &85.8 & 81.7 & 60.9 & {41.0} & {61.2}  \\
      \midrule
     SIFT, IJCV04 \cite{david2004distinctive}  & 17.2 &79.9 &90.3 & 47.4 & 34.1 & 57.3  \\
   \rowcolor[gray]{0.85}  SIFT~\cite{david2004distinctive} + GAMorph &12.5 \textcolor{blue}{(-4.7)} &89.5 \textcolor{blue}{(+9.6)} & 91.0 \textcolor{blue}{(+0.7)} & 56.8 \textcolor{blue}{(+9.4)}&42.1 \textcolor{blue}{(+8.0)}&63.3 \textcolor{blue}{(+6.0)}	 \\     
     SuperPoint, CVPRW2018  \cite{detone2018superpoint} & 11.8 &94.8 & 89.3 & 68.0 & 43.4  & 66.9 \\
    \rowcolor[gray]{0.85} SuperPoint~\cite{detone2018superpoint} + GAMorph &6.2 \textcolor{blue}{(-5.6)} &97.0 \textcolor{blue}{(+2.2)}&89.7 \textcolor{blue}{(+0.4)} & 71.7 \textcolor{blue}{(+3.7)} & 55.0 \textcolor{blue}{(+11.6)} & 72.2 \textcolor{blue}{(+5.3)}  \\ 
     SuperRetina, ECCV2022 \cite{liu2022semi} & {5.2} &{98.5} & 94.0 & {78.3} & 54.2 & 75.5  \\
     \rowcolor[gray]{0.85}  SuperRetina~\cite{liu2022semi} + GAMorph   &4.5 \textcolor{blue}{(-0.7)} &99.2 \textcolor{blue}{(+0.7)}& 94.4 \textcolor{blue}{(+0.4)} & 80.6 \textcolor{blue}{(+2.3)} & 65.9 \textcolor{blue}{(+11.7)} & 80.3 \textcolor{blue}{(+4.8)} \\ 
       GeoFormer, ICCV2023 \cite{liu2023geometrized} & {5.2} &{98.5} & 94.4 & 76.6 & {55.9} & {75.6}  \\
      \rowcolor[gray]{0.85} GeoFormer~\cite{liu2023geometrized} + GAMorph  & \textbf{4.5} \textcolor{blue}{(-0.7)}&\textbf{99.2} \textcolor{blue}{(+0.7)}& \textbf{95.8} \textcolor{blue}{(+1.4)} & \textbf{80.6} \textcolor{blue}{(+4.0)} & \textbf{66.3} \textcolor{blue}{(+10.4)} & \textbf{80.9} \textcolor{blue}{(+5.3)} \\ 
     \bottomrule
     \end{tabular}
     \label{tab:fire_results}
\end{table*}

%% file: tables/flori_result.tex
\begin{table}[t]
\centering
\caption{{Quantitative comparison of the proposed HybridRetina with  state-of-the-art methods on the FLoRI21 dataset~\cite{ding2021flori21}. We present the results of GAMorph with different keypoint detectors.The performance improvements of our method over the baseline are marked in blue.
}}
\resizebox{1.0\linewidth}{!}{%
\begin{tabular}{lccc}
    \toprule
      {Methods}  & RMSE $\downarrow$& Accept [\%] $\uparrow$& AUC [\%] $\uparrow$\\
    \midrule
    REMPE~\cite{hernandez2020rempe} &35.9  &46.7 &56.8 \\
    SIFT~\cite{david2004distinctive} &44.8  &33.3 &48.2    \\
    \rowcolor[gray]{0.85}SIFT~\cite{staal2004ridge} + GAMorph  & 38.7 \textcolor{blue}{(-6.1)} &40.0 \textcolor{blue}{(+6.7)} &54.1 \textcolor{blue}{(+5.9)}     \\
    SuperPoint~\cite{detone2018superpoint}	&23.7  &73.3 &68.3  	 \\
    \rowcolor[gray]{0.85}SuperPoint~\cite{staal2004ridge} + GAMorph  &19.2 \textcolor{blue}{(-4.5)} &73.3 \textcolor{blue}{(+0.0)}  &74.8 \textcolor{blue}{(+6.5)}     \\
    SuperRetina~\cite{liu2022semi}	 &18.0  &80.0 &77.1  	  	 \\
    \rowcolor[gray]{0.85}SuperRetina~\cite{staal2004ridge} + GAMorph &{13.1} \textcolor{blue}{(-4.9)} &{86.7} \textcolor{blue}{(+6.7)} &{84.3} \textcolor{blue}{(+7.2)}     \\
    GeoFormer~\cite{liu2023geometrized}	 &17.2  &80.0 &78.2	  	 \\
    \rowcolor[gray]{0.85}GeoFormer~\cite{liu2023geometrized} + GAMorph &\textbf{12.5} \textcolor{blue}{(-4.7)} &\textbf{86.7} \textcolor{blue}{(+6.7)} & \textbf{84.8} \textcolor{blue}{(+6.6)}     \\
       \bottomrule
     \end{tabular}
}
\label{tab:flori_test}
\end{table}

%% file: tables/ablation_component.tex
\begin{table}[t]
\centering
\caption{{Ablation study on key components of GAMorph. These experiments use SuperRetina as the keypoint detector and adopt the AUC [\%] for FIRE and FLORI21 datasets.
}}
\begin{tabular}{@{}cccccc@{}}
    \toprule
    Baseline &PAG &PPG &EAM  & FIRE & FLORI21\\
    \midrule
    \checkmark & & & & 75.9  & 77.9     \\
    \checkmark &\checkmark & & & 77.1   & 79.6  \\
    \checkmark &\checkmark &\checkmark &	& 78.7  & 82.1 \\
    \checkmark &\checkmark &\checkmark &\checkmark	 & 80.3  & 84.3   \\
       \bottomrule
     \end{tabular}
\label{tab:ablation_componet}
\end{table}

%% file: tables/ablation_deformable.tex
\begin{table}[t]
\centering
\caption{{Ablation study on using different deformable methods. The baseline directly uses the global matching results obtained from SuperRetina, while the other deformable methods are applied to the images after global registration.
}}
\begin{tabular}{@{}lcc@{}}
    \toprule
      {Methods}  & FIRE & FLORI21 \\
    \midrule
     SuperRetina, ECCV2022~\cite{liu2022semi} & 75.5  & 77.1  \\
     SuperRetina + Diff Demons, NeuroImage2009~\cite{vercauteren2009diffeomorphic} & 76.1  & 79.0  \\
     SuperRetina + VoxelMorph, TMI2019~\cite{balakrishnan2019voxelmorph} & 76.4  & 79.8  \\
     SuperRetina + AC-RegNet, Neural Netw2020~\cite{mansilla2020learning}   & 75.8  & 78.6  \\
     SuperRetina + Transmorph~\cite{chen2022transmorph}, MIA2022 & 77.4  & 81.2  \\
     SuperRetina + GAMorph (proposed) & \textbf{80.3}   & \textbf{84.3}  \\
       \bottomrule
     \end{tabular}
\label{tab:ablation_deform}
\end{table}